# Learning with partially separable data

Aida Khozaei[1], Hadi Moradi[1, 2], Reshad Hosseini[1]

[1] School of Electrical and Computer Engineering, University of Tehran, Tehran, Iran.

{a.khozaee, moradih, reshad.hosseini}@ut.ac.ir

[2] Adjunct Research Professor, Intelligent Systems Research Institute, SKKU, Suwon, South Korea

# Abstract

There are partially separable data types that make classification tasks very hard. In other words, only parts of the data are informative meaning that looking at the rest of the data would not give any distinguishable hint for classification. In this situation, the typical assumption of having the whole labeled data as an informative unit set for classification does not work. Consequently, typical classification methods with the mentioned assumption fail in such a situation. In this study, we propose a framework for the classification of partially separable data types that are not classifiable using typical methods. An algorithm based on the framework is proposed that tries to detect separable subgroups of the data using an iterative clustering approach. Then the detected subgroups are used in the classification process. The proposed approach was tested on a real dataset for autism screening and showed its capability by distinguishing children with autism from normal ones, while the other methods failed to do so.

# Introduction

The basic approaches in classification assume there are classes of instances with specific features used to distinguish these classes from each other. However, there are cases in which a subset of a class (a subgroup of instances) is discriminable or in an available feature space that is not a full representative of the whole class, it cannot be used to fully separate the class. For example, tip toe walking is a feature common in approximately 25% of children with Autism Spectrum Disorder (ASD) [1], i.e. a subgroup of children with ASD. In routine diagnosis procedures for detecting children with ASD, this feature is considered in parallel with other known ASD behavioral features. If walking (gait) data was the only data used for ASD diagnosis, about 75% of children with ASD would have not been distinguished. Similarly, there is a group of ASD children, not all of them, who have flapping feature [2]. In other words, there is a subgroup of a positive class, i.e. ASD class in this example, that is informative and can be distinguished in a feature space.

To clarify the mentioned situation, Figure 1 shows a visualization of a synthetic two-class data represented in a two-feature space (F1 and F2). As it is shown, only a subset of class circle is distinguishable in the present feature space. A subgroup, which includes instances with a black dot at their center is separable using the vertical line while the horizontal line separates another subgroup of the class circle, indicated by jagged circles.

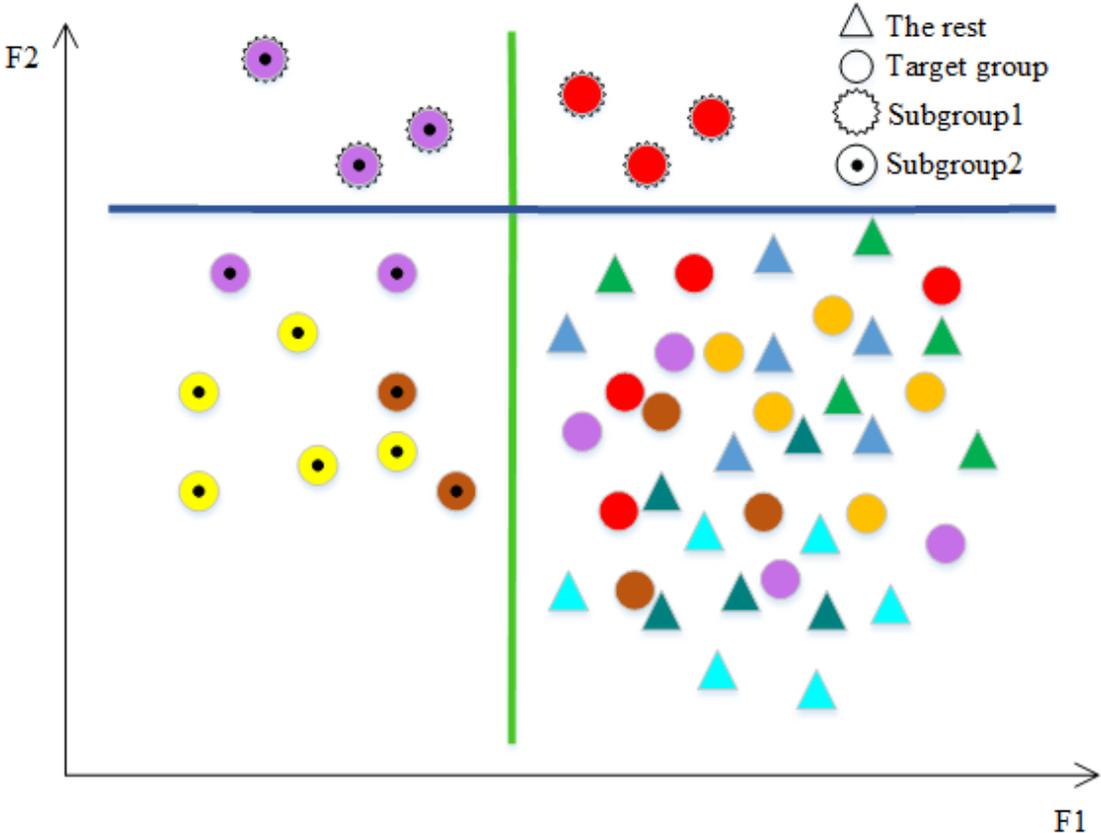

Figure 1: The instances of each subject of the positive class, i.e. the circles, and the negative class, i.e. the triangles, are shown using uniquely identifiable colors. There are two subgroups of the positive class

marked with black dots or with jagged surroundings. These subgroups can be distinguished using the vertical and horizontal lines respectively.

If we match the tip toe walkers example to the illustrated synthetic data example the circles with a black dot are the data from tip toe walks. Similarly, let's assume jagged circles are a subgroup of ASD class that represents flapping behavior.

As it is clear in the given example, the separable instances are not outliers or noisy labeled instances, despite being a few instances, and they should be considered in the classification. Consequently, using ordinary supervised Machine Learning (ML) methods [3], which assume all the data points are informative, the two classes are not separable. To address this problem partially, Multi-Instance Learning (MIL) [4] was introduced to consider cases in which a subset of instances of a subject in the target class are separable from the other classes. However, in the MIL approaches it is assumed that all the subjects (bags) in a class are distinguishable using a given set of features. However, not all subjects can be distinguished using the feature space in all data sets like tip toe walking or flapping examples. This problem can be shown in Fig. 1 with the orange-colored instances, which belong to a subject, that have no differentiable instance at all using the available feature space (F1 and F2).

Given the above shortcomings in the available methods, in this paper we introduce a framework that can handle partially separable data, similar to the situation given in Fig. 1. Actually, this framework was invented since the data from the cry sounds of children with ASD could not be classified using available methods [5]. In this paper, the general mathematical basis of the framework, an improvement to the previously proposed algorithm, and its expansion/usage to other problems is discussed. The proposed framework has been implemented and tested on a real datasets and the results show its capability to handling the discussed data specific situation.

# Related works

There are various machine learning methods with the assumption of having a dataset that is totally informative. Considering this assumption, many studies develop classifiers based on their whole available data [6-8]. On the other hand, there are cases with uninformative or misleading data among the others. Outliers and noisy labeled data are the two kinds of such problems that motivated many studies to develop methods in order to handle the misleading behavior of those uninformative parts [9, 10]. Also, there are studies that try to avoid overlapping data during training a classifier [11, 12]. Figure 2. Shows the above mentioned cases in a diagram. Under the category of the methods with uninformative data that includes noisy labeled, outliers and overlapping data, we introduce a subcategory of partially separable data discussed in Introduction. The inseparability in the data may be referred to inseparable instances or bags in ordinary ML or MIL problems, respectively. To the best of our knowledge there is no study in the literature that consider this assumption for a data type.

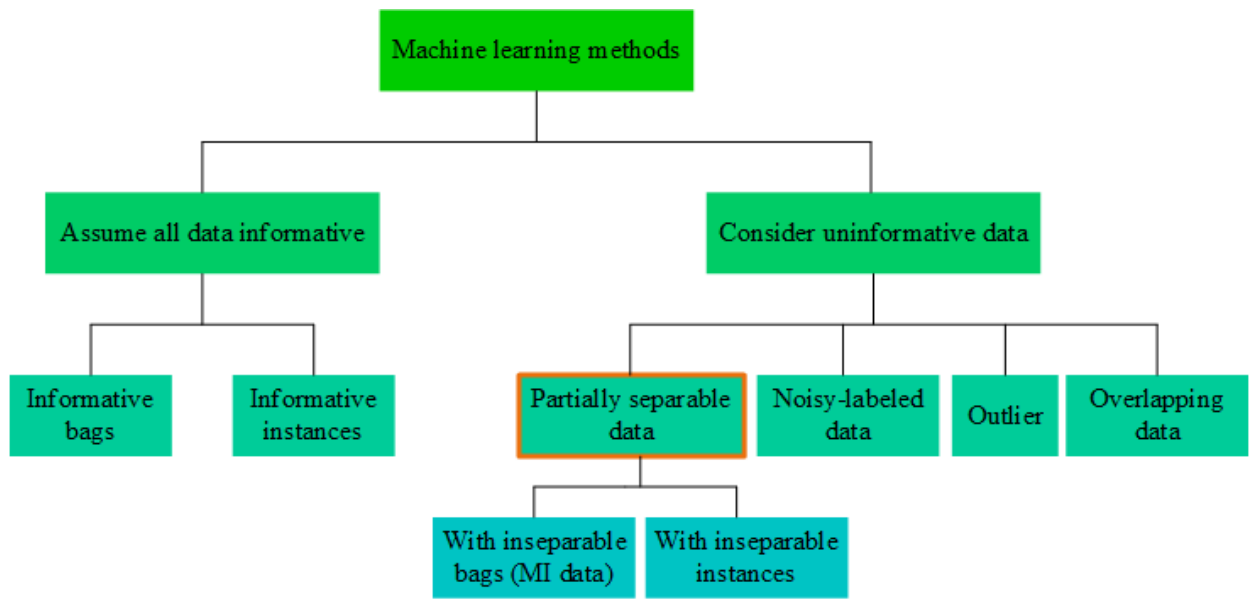

Figure 2. A diagram of methods for different data types based on having fully informative data or not. Partially separable data have parts of data with no specific information for separating their class.

In many medical applications, there are examples of asking the experts to locate the informative parts of the data. Xu, et al. [13] used images with obvious cancerous tissues in training phase as positive labeled images. Sudharshan, et al. [14] used images of benign and malignant breast tumors which were generated from breast tissue biopsy slides labeled by pathologists [15]. They asked experts to label positive and negative images and assumed each image as a bag to use MIL methods. Taban, et al. [16] used only the instances of walking on tiptoe to get rid of the problem of having indistinguishable instances, i.e. normal walking steps. All these works rely on human opinion to avoid using inseparable data. This is the case for many similar studies in computer-aided medical diagnosis or classification [17, 18]. MIL methods that use bags of instances, assume that there is at least one instance (or a specific portion of instances) separable in each positive bag [4, 19]. So, they also assume the data, in bag level, is separable [20].

Having inseparable data makes the classification results weak. An approach to strength the results is considering all data to be informative, use them in training phase and then tune hyperparameters of the classifier, e.g. thresholding on the accuracy, to solve the problem caused by the uninformative data. Pokorny, et al. [21] used audio recording segments of infants' vocalizations as instances of children with autism and their normal peers. They trained several classifiers on all labeled instances and used two approaches of majority voting and setting a hyperparameter which is a threshold based on the best accuracy achieved in a validation set, for labeling each subject. Using an adaptive threshold hyperparameter improved the results compared to majority voting (threshold of 50%). This kind of approach to handle uninformative data may not be suitable when the uninformative data is not near the boundary. For example, the reported results of [21] show that there is a subject with no separable instance during the cross validation, using the applied classifier. So there are instances of a subject that they are completely inseparable using the classifier and that affects setting the threshold. [5] also shows that using usual classification methods on a data with inseparable data does not have appropriate results, even by giving the best chance by selecting the best threshold even based on the test results.

[11] proposed a method to avoid focusing on hard-to-learn samples by removing instances that were not correctly classified by a 'perfect' Bayesian model. [12] provided a label-confidence based boosting method to get immune to the label noise and overfitting problems. These studies try to avoid classifiers form engaging with overlapped samples and focus on others. This overlapping can come from noise in the data or noise of the classifier. These studies' assumption of having reliable initial guesses by the classifiers to be used for assigning confidence or avoidance criterion for instances, relies on a basic assumption of having the majority of instances with reliable information for the classifier. So, in cases there are a significant portion of instances in overlapped area the average confidence is not reliable at all and getting rid of the instances that mislead the classifiers at the first stages is necessary. In addition, when overlapped instances of, for example, two opponent classes fall in similar distribution, relying on the confidence is not meaningful.

Noisy labeled data refers to any data with incorrect labels [9]. So, noisy labeled instances can decrease a classifier performance. There are researches that propose methods to handle noisy labeled data [22-24]. To handle noisy labeled data which is made by wrong labeled instances due to the complexity of data labeling even for experts, big data with messy user tags, or any other reason which is a natural result of data collection process, there is an assumption of having wrong labeled instances [9]. For example, [25] that proposed a self-sampling method to decrease the effect of noisy labeled data, changed the labels of several instances in its experimental tests to simulate the problem of having noisy instances. Although these methods consider removing noisy labeled data that is an inseparable portion of a class of data and use other parts, the assumption behind their developments is different from what is focused in this study. Having inseparable parts in a class of data that has similar distribution to its opponent class is not considered in developing such methods. In addition, the inseparable parts in noisy labeled data is not considered to be a majority portion of the data while this consideration is natural when assuming a class with partially separable data has even a minority of separable data [5].

Outliers usually composes a small percentage of data, which comes from a different distribution. The task of outlier detection is to find a minor fraction of data samples, which is far different from the most samples [26, 27]. Outlier detection methods have been usually developed based on the assumption of having a fraction of lower than about 5% as outlier, to detect such instances [28]. In methods that can handle outliers, there is no assumption of having inseparable data that misleads the classifiers [9, 10].

## Partitioning data to strengthen classifiers

There are studies that propose methods that make classifiers, which are weak for a data type, a stronger one by considering subsets or partitions of instances instead of all. This topic may not be directly related to our method but they can be considered in a view point of focusing in parts of data to make classifiers stronger in separating data.

There are methods that try to create a strong classifier by ensembling several weak classifiers. Each weak classifier can work on a part of data, and therefore other data can be considered to be inseparable for that classifier. For example, cascading methods try to have positive instances classified correctly at very early stages while not all negative ones may be classified correctly. In each stage, they may remove obvious or correctly classified negative instances of the previous stages for training [29]. Boosting algorithms try to upweight instances that are wrongly classified in early weak classifiers [30] or select and use hard-to-be-classified negative instances that are more informative than others [31]. While these mentioned approaches groups instances to improve classification tasks, they have no assumption of having inseparable data. Again, the assumption of having all data separable and forcing the classifier to separate them may mislead the classifier.

From another point of view, there are studies try to improve their classifiers performances by dividing the data to regions based on some assumptions. For example [32] assumes that the data comes from multi distributions and try to finds them by using nearest neighbor clustering that leads to divide the data into subclasses before using discriminant analysis. Mixture of experts methods, use few classifiers as experts in different regions that are trained separately on the whole data and a gating network is involved in assigning each classifier to its space of specialization [33]. In these methods, it is assumed that the data can be classified correctly in each region and therefore they cannot be used to solve the problem of partially separable data that have naturally inseparable parts.

# The proposed framework

In this section a framework suitable for supervised classification with partially separable instances is proposed.

In supervised classification problems given a set $X$ consists of data points (instances) in a feature space, and a set $Y$ including labels, the goal is to estimate the following function $f$:

$$f: X \rightarrow Y$$

Considering $\{X, Y\}$ as the whole set of instances and their labels in the training set, $\{X_p, Y_p\}$ and $\{X_n, Y_n\}$ are the set related to positive and negative class instances relatively.

As shown in figure 1, for a case of having a positive class, i.e. the circles, with partially separable data, a hypothetical classifier can separate a set of the instances of circles. In other words, $X_p$ can be partitioned using a variable $Z$, into several subgroups that is hidden. This distribution partitions the target data of $X_p$ into $Z_1, Z_2, \ldots, Z_i$. Since the data are partially separable, only for some $Z_k \in Z_1, Z_2, \ldots, Z_i$, the mutual information $I(\{x_p|Z_k, x_n\}; y)$ is high.

So, one way to solve the partially separable data problem is to find the set of the separable partitions that is:

$$Z_s = \{Z_i | I(\{x_p|Z_i, x_n\}; y) \text{ is high}\}$$

Since the mentioned partitions are not visible, an assumption is necessary to make the problem solvable. It is assumed that the data of separable partitions are close to each other and far from other data in the space (they are clustered). Using any method based on the mentioned assumption to find the subspaces related to the mentioned informative (separable) subgroups, we will have the set $Z_s$ and then the instances of each partition will be used versus the instances of the negative class in training a classifier as it is shown in figure 3. Classifiers are combined with each other in a way that each instance detected by any of them will be labeled as positive instance. In other words, each instance detected as a subgroup instance belong to the positive class.

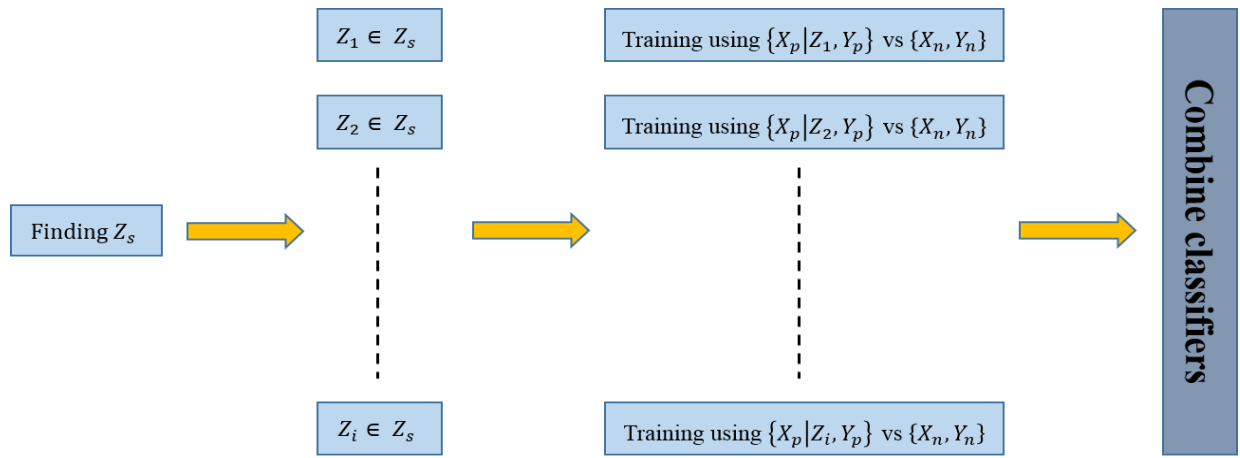

Figure 3. The steps for training classifiers using separable partitions of positive class data, based on hidden separable partitions $Z_1, Z_2, \ldots, Z_i \in Z_s$, versus negative class data.

## An algorithm based on the proposed framework

Here, we propose an algorithm which is an extension of the algorithm given in our previous [5] work. As mentioned earlier, it is assumed that the data can be clustered, so clustering can be used to find separable subgroups of positive instances, that we call them exclusive clusters. Then, classification can be used to train classifiers based on the exclusive clusters. It should be mentioned that we may allow an exclusive cluster to include few negative instances to handle noise, outliers, or overlaps. Therefore, the framework algorithm will get robust to noisy, overlapped or outlier data. Algorithm 1 shows the pseudocode for determining exclusive clusters and training classifiers.

| Algorithm 1. The training phase of the proposed algorithm |
|---|
| $P$: set of all positive instances |
| $N$: set of all the negative instances |
| $F$: set of all classifiers |
| $\rho$: threshold for the number of instances in a cluster |
| $s$: the number of minimum positive instances needed in a cluster to be able to train a classifier for it |
| $t$: threshold for the accepted portion of the negative instances in an exclusive cluster |
| $st$: threshold for the sensitivity of a classifier |

| $C_j$: the $j^{th}$ cluster | ;in the beginning the whole data is assumed as one cluster with $j = 1$ |
|---|---|
| $P_j$: the positive instances in $j^{th}$ cluster | |
| $N_j$: the negative instances in $j^{th}$ cluster | |
| $n$: number of clusters | |
| $F = \emptyset, EC = \emptyset, n = 1,$ | |

| | | | | | |
|---|---|---|---|---|---|
| 1: | While $\exists j \, |C_j| > \rho$ or n =1 | | | | ; while there is a cluster bigger than a threshold or n =1 |
| 2: | | $n = n + 1$ | | | ; increase the number of clusters |
| 3: | | Cluster the $P + N$ into $n$ clusters $C_j, j = 1, ..., n$ | | | |
| 4: | | $EC = \{C_j | \, |N_j|/|P_j| < t \text{ and } |P_j| > s\}$ | | | ; the set of exclusive clusters with acceptable size and tolerating a maximum portion of negative instances |
| 5: | | If $EC \neq \emptyset$ | | | ; check if there is any acceptable exclusive cluster |
| 6: | | | For all $C_j$ in $EC$ | | |
| 7: | | | | Train a classifier using positive labeled pos $\in C_j - N_j$ and negative labeled neg $\in N$ | |
| 8: | | | | If classifier sensitivity is higher than $st$ | |
| 9: | | | | | Add the classifier to $F$ |
| 10: | | | | | $P = P - P_j$ ; remove the instances of the exclusive clusters from positive group instances |
| 11: | | | | | $n = 1$ |

Algorithm 1 is based on the loop in line 1 that starts to have data clustered, initiating with two clusters (Line 2). Then the number of clusters is increased until achieving at least one exclusive cluster or no exclusive cluster, with a predefined minimum size, can be found. The maximum threshold of having negative instances in an exclusive cluster is checked in line 4. In line 7 the positive instances of a detected exclusive cluster are used to train a classifier. In lines 8 and 9, the classifiers with a sensitivity higher than a given threshold are added to the set of trained classifiers. Then, in line 10, the positive instances of such exclusive clusters are removed for the data set. In line 11, the numbers of clusters are set to 1 and the loop restarts. The loop stops when the number of instances in each cluster is less than a threshold.

After having all classifiers trained to classify exclusive clusters, an ensemble approach can be used to decide the result of classification for a given input.

# An application

As mentioned earlier, the proposed framework was invented due to the need to classify cry sounds of children with ASD, which could not be classified with the available methods. Specifically, different available machine learning methods were tested and finally Radial Basis Support Vector Machine (RBF-SVM) were selected with better results than the others. Despite the better performance of the trained RBF-SVM compared to the rest of methods, it had poor results, i.e. very bad sensitivity, on the test set of 28 boys, including 14 children with ASD and 14 TD children (Table 1). In this classifier the majority pooling decision making was used on the results of the instances on each participant. In contrast, the proposed algorithm had a much higher sensitivity of 85.71% compared to the 35.71% sensitivity of the RBF-SVM algorithm. It should be mentioned that in the proposed algorithm a participant was labeled as ASD if at least one of the trained classifiers tagged one of the participant's instances as ASD.

We even tried the best-chance threshold pooling, which is a threshold-based pooling in which a threshold is selected based on the best accuracy on the test set. Even using this best-chance pooling could not produce any better results than the proposed algorithm. The results are shown in Table 1.

Table 1: Results of using SVM as the best classifier among others in usual training manner vs using our proposed method. MP stands for the majority pooling and BP stands for the best-chance pooling

| Specificity | | | Sensitivity | | |
|---|---|---|---|---|---|
| SVM | | SSI | SVM | | SSI |
| MP | BP | | MP | BP | |
| 100 | 85.71 | 100 | 35.71 | 71.42 | 85.71 |

# Discussion and Conclusion

In this paper a simple and practical framework for classification of data with partially separable parts is proposed. The proposed framework does not have the typical assumption that the separating property is valid across all instances which may mislead various methods in pattern recognition.

The proposed framework is also applicable for classification MI data types. In fact, using separable instances in a general manner makes it also applicable to typical MI data sets, which have bags of instances with several separable instances among all, as well as MI data sets with inseparable bags. An algorithm based on the proposed framework is presented and its application on a dataset of the cry sounds of children with autism, which has inseparable bags, has been shown with great results.

A modification in the proposed framework with the consideration of having separable instances in the negative class as well as the positive class, can lead to have a framework to reject inseparable instances. In other words, in such modification we assume there is a portion of instances of both classes that comes from a same distribution while there are other instances in each class from different distributions that makes them separable. This way, a preprocessing phase for removing the common inseparable instances could be done as a rejection phase before the training phase.

As an extension to the proposed framework, a feature selection phase for each subgroup can be incorporated. Actually, this framework helps finding the features that are important in detecting the subgroups of a class. Assuming $f_1, f_2, \ldots, f_d$ are the features available in the d-dimensional problem. Feature selection should be adopted for each subgroup. For each $Z_i \subset Z_s$ we can use a feature selection approach, for example maximizing $\sum_{j=1}^{d} I(f_j; Y)$, to find related features for each subgroup. As it is shown in our previous work [5], after finding the subgroups, features related to each group were selected and it makes the classification possible where the training data set was small relative to the original feature space.

References:

[1] W. J. Barrow, M. Jaworski, and P. J. Accardo, "Persistent toe walking in autism," *Journal of Child Neurology,* vol. 26, pp. 619-621, 2011.

[2] Centers for Disease Control and Prevention. (2019, 29 Feb). *Signs and Symptoms of Autism Spectrum Disorders*. Available: https://www.cdc.gov/ncbddd/autism/signs.html

[3] S. Theodoridis and K. Koutroumbas, *Pattern recognition*: Elsevier, 2003.

[4] J. Foulds and E. Frank, "A review of multi-instance learning assumptions," *The Knowledge Engineering Review,* vol. 25, pp. 1-25, 2010.

[5] A. Khozaei, H. Moradi, R. Hosseini, H. Pouretemad, and B. Eskandari, "Early screening of autism spectrum disorder using cry features," *PLOS ONE,* vol. 15, p. e0241690, 2020.

[6] J. Qin, W. Pan, X. Xiang, Y. Tan, and G. Hou, "A biological image classification method based on improved CNN," *Ecological Informatics,* vol. 58, p. 101093, 2020/07/01/ 2020.

[7] R. C. Deo, "Machine Learning in Medicine," *Circulation,* vol. 132, pp. 1920-1930, 2015.

[8] E. R. Davies, "Chapter 13 - Basic classification concepts," in *Computer Vision (Fifth Edition)*, E. R. Davies, Ed., ed: Academic Press, 2018, pp. 365-398.

[9] B. Frénay and M. Verleysen, "Classification in the presence of label noise: a survey," *IEEE transactions on neural networks and learning systems,* vol. 25, pp. 845-869, 2013.

[10] A. Smiti, "A critical overview of outlier detection methods," *Computer Science Review,* vol. 38, p. 100306, 2020.

[11] A. Vezhnevets and O. Barinova, "Avoiding Boosting Overfitting by Removing Confusing Samples," Berlin, Heidelberg, 2007, pp. 430-441.

[12] Z. Xiao, Z. Luo, B. Zhong, and X. Dang, "Robust and Efficient Boosting Method Using the Conditional Risk," *IEEE Transactions on Neural Networks and Learning Systems,* vol. 29, pp. 3069-3083, 2018.

[13] Y. Xu, J.-Y. Zhu, I. Eric, C. Chang, M. Lai, and Z. Tu, "Weakly supervised histopathology cancer image segmentation and classification," *Medical image analysis,* vol. 18, pp. 591-604, 2014.

[14] P. J. Sudharshan, C. Petitjean, F. Spanhol, L. E. Oliveira, L. Heutte, and P. Honeine, "Multiple instance learning for histopathological breast cancer image classification," *Expert Systems with Applications,* vol. 117, pp. 103-111, 2019.

[15] F. A. Spanhol, L. S. Oliveira, C. Petitjean, and L. Heutte, "A Dataset for Breast Cancer Histopathological Image Classification," *IEEE Transactions on Biomedical Engineering,* vol. 63, pp. 1455-1462, 2016.

[16] R. Taban, A. Parsa, and H. Moradi, "Tip-Toe Walking Detection Using CPG Parameters from Skeleton Data Gathered by Kinect," in *International Conference on Ubiquitous Computing and Ambient Intelligence*, 2017, pp. 287-298.

[17] Ü. Budak, Z. Cömert, Z. N. Rashid, A. Şengür, and M. Çıbuk, "Computer-aided diagnosis system combining FCN and Bi-LSTM model for efficient breast cancer detection from histopathological images," *Applied Soft Computing,* vol. 85, p. 105765, 2019.

[18] P. Moeskops, M. A. Viergever, A. M. Mendrik, L. S. De Vries, M. J. Benders, and I. Išgum, "Automatic segmentation of MR brain images with a convolutional neural network," *IEEE transactions on medical imaging,* vol. 35, pp. 1252-1261, 2016.

[19] J. Amores, "Multiple instance classification: Review, taxonomy and comparative study," *Artificial Intelligence,* vol. 201, pp. 81-105, 2013.

[20] M.-A. Carbonneau, V. Cheplygina, E. Granger, and G. Gagnon, "Multiple instance learning: A survey of problem characteristics and applications," *Pattern Recognition,* vol. 77, pp. 329-353, 2018/05/01/ 2018.

[21] F. B. Pokorny, B. W. Schuller, P. B. Marschik, R. Brueckner, P. Nyström, N. Cummins*, et al.*, "Earlier Identification of Children with Autism Spectrum Disorder: An Automatic Vocalisation-Based Approach," in *INTERSPEECH*, 2017, pp. 309-313.

[22] T. Liu and D. Tao, "Classification with noisy labels by importance reweighting," *IEEE Transactions on pattern analysis and machine intelligence,* vol. 38, pp. 447-461, 2015.

[23] T. Leung, Y. Song, and J. Zhang, "Handling label noise in video classification via multiple instance learning," in *2011 International Conference on Computer Vision*, 2011, pp. 2056-2063.

[24] W. Feng, Y. Quan, and G. Dauphin, "Label Noise Cleaning with an Adaptive Ensemble Method Based on Noise Detection Metric," *Sensors,* vol. 20, p. 6718, 2020.

[25] X. Liu, S. Luo, and L. Pan, "Robust boosting via self-sampling," *Knowledge-Based Systems,* vol. 193, p. 105424, 2020.

[26] Z. Wang, Z. Zhao, S. Weng, and C. Zhang, "Incremental multiple instance outlier detection," *Neural Computing and Applications,* vol. 26, pp. 957-968, 2015.

[27] O. Wu, J. Gao, W. Hu, B. Li, and M. Zhu, "Identifying multi-instance outliers," in *Proceedings of the 2010 SIAM International Conference on Data Mining*, 2010, pp. 430-441.

[28] R. Domingues, M. Filippone, P. Michiardi, and J. Zouaoui, "A comparative evaluation of outlier detection algorithms: Experiments and analyses," *Pattern Recognition,* vol. 74, pp. 406-421, 2018.

[29] W. Ouyang, K. Wang, X. Zhu, and X. Wang, "Chained cascade network for object detection," in *Proceedings of the IEEE International Conference on Computer Vision*, 2017, pp. 1938-1946.

[30] D.-S. Cao, Q.-S. Xu, Y.-Z. Liang, L.-X. Zhang, and H.-D. Li, "The boosting: A new idea of building models," *Chemometrics and Intelligent Laboratory Systems,* vol. 100, pp. 1-11, 2010.

[31] J. Robinson, C.-Y. Chuang, S. Sra, and S. Jegelka, "Contrastive learning with hard negative samples," *arXiv preprint arXiv:2010.04592,* 2020.

[32] M. Zhu and A. M. Martinez, "Subclass discriminant analysis," *IEEE transactions on pattern analysis and machine intelligence,* vol. 28, pp. 1274-1286, 2006.

[33] S. E. Yuksel, J. N. Wilson, and P. D. Gader, "Twenty years of mixture of experts," *IEEE transactions on neural networks and learning systems,* vol. 23, pp. 1177-1193, 2012.